# Design and Development of a Framework For Stroke-Based Handwritten Gujarati Font Generation

**Preeti P. Bhatt** [0000-0003-4668-9061], **Jitendra V. Nasriwala**[0000-0002-5585-1668] **and Rakesh R. Savant**
[0000-0003-3497-8026]

*Babu Madhav Institute of Information Technology, Uka Tarsadia University, Bardoli, India*

*Email address: preeti.bhatt@utu.ac.in, jvnasriwala@utu.ac.in, rakesh.savant@utu.ac.in,*

**Abstract:**

Handwritten font generation is important for preserving cultural heritage and creating personalized designs. It adds an authentic and expressive touch to printed materials, making them visually appealing and establishing a stronger connection with the audience. This paper aims to design a framework for generating handwritten fonts in the Gujarati script, mimicking the variation of human handwriting. The proposed font generation model consists of a learning phase and a generation phase. In the learning phase, Gujarati scripts are analyzed, and rules for designing each character are formulated. This ruleset involves the concatenation of strokes in a stroke-based manner, ensuring visual consistency in the resulting glyphs. The generation phase involves the user providing a small subset of characters, and the system automatically generates the remaining character glyphs based on extracted strokes and learned rules, resulting in handwritten Gujarati fonts. The resulting character glyphs are converted into an open-type font using the FontForge tool, making them compatible with any Gujarati editor. Both subjective and objective evaluations are conducted to assess the synthesized images and fonts. Subjective evaluation through user studies provides feedback on quality and visual appeal, achieving an overall accuracy of 84.84%. Notably, eleven characters demonstrated a success ratio above 90%. Objective evaluation using an existing recognition system achieves an overall accuracy of 84.28% in OCR evaluation. Notably, fifteen characters had a success ratio of 80% or higher.

## 1. INTRODUCTION

Text-based communication is rapidly gaining prominence in education, business, and personalization, seamlessly connecting people across the globe with zero cost and minimal time constraints. Moreover, the use of computer fonts has become universal in facilitating textual communication. In the present era, text is ubiquitous, appearing in emails, messages, books, posters, text messages, computers, and a myriad of other mediums. This text is often presented in a plethora of computerized fonts meticulously crafted by font engineers. Despite the burgeoning number of fonts generated by these designers, they still fall short of satisfying the demand for personalized fonts capable of conveying the writer's emotions and identity. The solution to this challenge lies in adopting one's own handwriting style, which not only enhances reader comfort but also adds a personal touch. Achieving this level of personalization entails the process of handwritten font synthesis. Handwriting synthesis is the transformation of input data into handwritten text, faithfully replicating the user's unique writing style. This synthesis is not only instrumental in infusing a personal touch but also finds utility in a diverse range of real-world applications, including the digitization of personal documents, writer identification, forgery detection, improving recognition systems, and the restoration of historical documents, among other uses [1]. Creating handwritten fonts that encapsulate key elements of the user's writing style is no easy feat, especially in languages with smaller character sets like English, which only comprise 26 letters. For example, numerous font development tools are readily available, including FontForge [2], Calligrapher, Font creator[3], font lab[4], and others. These tools demand input of a mere 26 letters, and they can promptly generate an entire font that adheres to the user's specifications.

The existing research in the field of font generation has primarily focused on languages such as Arabic[2]–[11], Chinese [12]–[27], Latin (English and Spanish) [28]–[37], Bangla[38]–[40], Japanese[41]–[43], Korean[44]–[46] and Indian[39], [47]–[49]. However, there has been limited work reported specifically for the Gujarati script, which highlights the need for further research in this area. Due to the unique and complex structure of the Gujarati script, existing resources developed for other scripts cannot be directly applied. Therefore, the proposed research on handwritten Gujarati font generation becomes crucial to address this gap.

Generating handwritten Gujarati fonts is challenging due to the script's complexity and variability in individual writing styles. Challenges include capturing individuality, dealing with a vast character set, limited datasets, character and size variations, document-to-document differences, and maintaining a reasonable recognition rate. These challenges require a deep understanding of Gujarati script and typographic design to create authentic synthesized fonts for practical use.

This article introduces a novel approach to the automatic generation of user wise handwritten Gujarati font libraries. The entire system is divided into two main phases: the learning phase and the generation phase.

In the learning phase, rules are established for generating each consonant. The standard handwritten Gujarati dataset from TDIL [25] is utilized, and the ruleset is designed based on features such as strokes' joining points, orientation, size, and occurrence. Distinct rules for individual classes are prepared through statistical analysis of each character sample from the dataset. In the generation phase, users are required to write a small set of characters on a blank sheet of paper. Subsequently, the system autonomously generates all other Gujarati consonants in the same style, creating a writer-dependent font library for Gujarati consonants. Using this model for synthesizing handwritten fonts proves to be highly convenient, as the writer only needs to provide a small, carefully selected subset (8% of samples) of characters on the blank page. Handwritten character samples are then imported into the system to segment individual characters. After acquiring image text samples, stroke extraction is performed for each segmented character, followed by stroke labeling. Once the strokes are properly labeled, the system generates the other characters by reusing those strokes and the rules prepared during the learning phase.

The rest of the document is structured as follows: In section 2, related work in the field of handwriting synthesis is presented. Following that, in section 3, the proposed methodology is described. In section 4, the results of the experiments are presented. Finally, the document concludes by discussing future work in the last section.

## 2. RELATED WORK

There are several applications of handwriting synthesis, such as Editable handwritten files with real-time inline correction, document digitization, historical documents repairing, font graphics, captcha generation, and many more[1]. Moreover, it can also be helpful to forensic examiners, the disabled, and researchers working on handwriting recognition systems[19]. Accordingly, there are two handwriting synthesis approaches Movement simulation and shape simulation.

The movement simulation approach uses neuromuscular hand movement to simulate handwriting. This approach considered writing skills, hand movement, and pen pressure as features that finally used to the curvature of handwritings. Their system focused on physical aspects rather than actual writing, which investigates handwriting ability[26]. The shape simulation contrary approach uses shape, curvature for simulating the writing. The shape simulation approach is classified into two phases; generation and concatenation. In the generation phase, the statistical information must be extracted from sufficient samples and imitate ultimately new models [8], [9], [45], [50]–[52], [28]–[33], [38], [41] . The concatenation approach is a fusion-based technique that blends two or more strokes to produce the new one [13], [20]–[27].

System using describes approach is used to synthesize handwriting with different scripts such as English, Chinese, Arabic, Indian. Significant work has been reported for Chinese writings, especially with the concatenation approach. [6] proposed a Chinese synthesis method to build the primary Chinese characters to generate standard Chinese characters. Their process used the affine transformation model to imitate the user and used the shape similarity method to evaluate the results. In [23], they designed an app for smartphones. Users need to write one by one Chinese characters. Then the app tracks every stroke from writing and finally synthesizes targeted Chinese characters from these extracted strokes using the concatenation approach.

Also, significant work is reported with the generation approach for the English script. [13] proposed an algorithm that imitates the input string in an author's handwriting. The system can simulate historical documents in the author's style. They used glyph centric approach with the learned parameter for line thickness, spacing, and pressure to produce handwriting that looks handmade. But they targeted only English characters. [12] presented an elegant and effective way to synthesize English handwriting in the user's writing style. They used essential features such as character glyph, size, slant, pressure, connection style, letter spacing, and cursive. Their model can produce personal handwriting with good visual quality. However, the study only supports English writing. In [17], two concatenation models are adopted to synthesize Arabic text from segmented characters. They aim to improve recognition performance significantly, not for imitating actual user writing

Handwriting generation for Indian scripts is a complex and significant research area, involving diverse styles and characteristics. In [47], deep learning and cultural knowledge are applied to tackle online handwriting generation challenges, paving the way for future advancements. [39] focuses on creating a synthesizer for Bengali and Devanagari scripts, successfully generating authentic signatures using the motor equivalence model. [48] introduces a novel approach for synthetic datasets in word recognition, achieving high accuracy for Indian scripts, thereby enhancing recognition systems. [49] presents a unique method for synthetic datasets, preserving patterns with added distortion, and improving recognition for Devanagari and Bengali scripts. Generating Gujarati handwriting presents challenges due to its complex structure and the limited research available. Resources developed for other scripts cannot be directly applied to the Gujarati script because of its unique features and distinct glyphs. However, insights gained from studying other Indian scripts can be beneficial. Specialized methods are required to accurately capture the nuances of Gujarati writing. The limited research in Indian scripts, particularly for Gujarati, has encouraged the development of the Handwritten Gujarati Font via stroke-based synthesis, as proposed in the next section.

## 3. FRAMEWORK FOR HANDWRITTEN GUJARATI FONT GENERATION (HGFG)

We intend to learn the user's handwriting style from the small characters and then automatically generate the handwritten font in the user style. We have proposed a handwritten Gujarati font generation model with mainly two phases: learning and generation. In the learning phase, the standard Gujarati characters' dataset was taken and the ruleset for each character to simulate the characters. In the generation phase, the user needs to write a small number of characters and simulate the glyph of other characters from the style and shape of inputted characters' strokes and ruleset defined in phase 1. Finally, an OpenType font is developed based on Gujarati scripts and Unicode. Figure 1 shows the complete model of handwritten Font synthesis.

### a. LEARNING PHASE

In the learning phase, we label all necessary information such as size, position, distance between disconnected components, height, width, joining point, number of endpoints from input characters and, targeted character set.

To illustrate the strokes utilized in the synthesis of the desired characters, Figure 2 is presented. This figure highlights two endpoints, labelled A and B, which serve as key reference points for the selected strokes. Through careful observation and analysis of the stroke patterns, it becomes possible to identify the specific strokes that need to be combined to generate the desired characters.

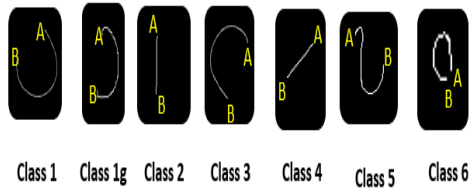

Figure 2 Strokes with class labels are considered to generate targeted characters

#### i. RULESET PREPARATION

The features of individual character classes have been extracted from the standard dataset of handwritten Gujarati characters [53] and then prepared ruleset, which helps to generate glyphs of the respective character. For developing a ruleset, we have fixed some standard features for all target characters and some variable features as per the requirement of characters. In standard features, we have considered the position of strokes, the occurrence of strokes, the strokes' size, and the Joining point. In variable features, we have considered the distance between two strokes and the appearance of strokes. The detailed description of features is shown in Table 1.

In certain characters like ક, ટ and ડ the joining points are simply the endpoints of the strokes, labelled as A or B. The joining points in these characters are easily identified. However, in some cases like ચ, વ, ધ, ધ etc., the joining points need to be determined through statistical analysis. For example, the character ચ consists of two strokes, the 'ta-shape' (ટ) and the 'line,' (ગ). So, at what specific point these two strokes join is decided by extracting the junction point from each sample of 'ચ', and the value is normalized by taking the percentage. Finally, choose the junction point by taking the mode of the normalized point. Similarly, another ruleset is derived by statistical analysis of the features of each targeted character. Based on the study, it has been observed that only one rule is required for the majority of characters, whereas, in characters ૐ, ચ, ઠ, ઢ, and સ, two rules have been identified as the statistical ratio are close with each other. In this case, we have generated two different shapes.

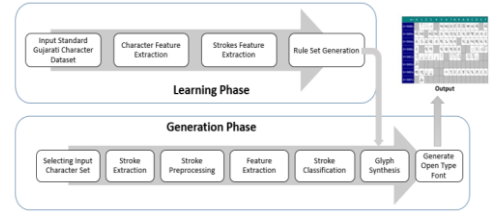

Figure 1 Framework for handwritten Gujarati font generation

Table 1 Features which are considered for preparing the ruleset with its description

| Features | Feature Type | Short Form | Description |
|---|---|---|---|
| Position | Common | POS | Position of stroke in specific characters |
| Occurrence | Common | OCC. | The number of times strokes occurred in characters. |
| Size | Common | S | Size of strokes: same or need to reform. |
| Joining Point | Common | JP. | The point at which two strokes combine. |
| Distance between two strokes | Variable | DS. | Distance needs to be kept between two strokes. |
| Appearance | Variable | AP. | The appearance of stroke: same or need to flip or rotate. |

### b. GENERATION PHASE

The core objective of this phase is to generate the glyph of targeted handwritten Gujarati characters from the small samples written by the user and the ruleset derived in the learning phase. This entire process is carried out by following substages.

#### i. SELECTING INPUT CHARACTER SET

To generate handwritten characters in the user's style, we must first analyze a few samples of their handwriting. The key question is determining the number of characters the user intends to write. To capture a comprehensive range of strokes, we collect a set of characters from the user. In this study, we focus on the character set ક, ખ, and ગ and utilizing six fundamental strokes. Our character generation process incorporates user input features, including the average character width, spacing between disconnected components, and other relevant information.

### ii. STROKE EXTRACTION

Our method, described in [54], extracts character strokes from thinned binary images by identifying endpoints and junction points. Endpoints, where a stroke starts and ends, are characterized by having only one connecting neighbour. We use eight structuring elements and a hit-or-miss transformation to detect them. A 3x3 structuring template is matched with pixel elements to set them to 1 if matched entirely, or 0 if not. Junction points occur when one pixel connects with more than two neighbours. We employ four T-Junction, seven Y-Junction, and seven Cross-Junction (3x3 structuring templates, +, ×) for junction point detection, applying a hit-or-miss morphological operation to each template for endpoint detection. Removing these points from the thinned binary skeleton results in disjoint images. We extract strokes from these disjoint images using a connected component approach. This process continues until each stroke has only two endpoints and no more junction points. The extracted strokes are depicted in Figure 3, showcasing the results of the stroke extraction process.

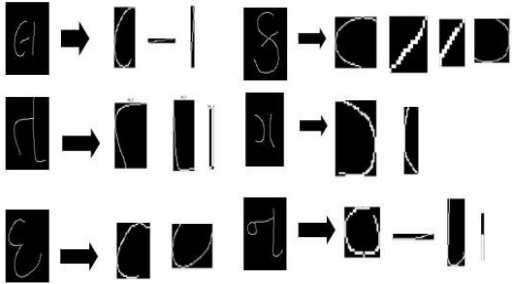

Figure 3 Result of the proposed method on a few characters in the form of stroke segmentation.

### iii. STROKE PREPROCESSING:

Each segmented stroke is resized to 28x28 and then subjected to binary conversion, grayscale, erosion, and dilation. Otsu's thresholding method is used for binary conversion. Thinning is performed using the Zhan Suen algorithm to create a one-pixel-wide thinned image. Finally, the image is padded by one pixel in each direction, resulting in a 30x30 image.

#### 1. ADAPTIVE THINNING

The ZhangSuen algorithm has shown improvements in thinning handwritten Gujarati characters. However, there are still instances where the resulting image is not consistently one pixel wide, which is necessary for further processing. To address this issue, we propose an enhanced thinning algorithm specifically designed for Gujarati handwritten strokes in [55]. In our approach, we leverage the characteristics of strokes, which are the smallest components of a character with two endpoints and no additional junction points. Taking this into consideration, we have developed a straightforward method using a 3x3 neighbourhood. To achieve a one-pixel width image, each 3x3 neighbourhood should contain fewer than four active pixels. If a neighbourhood has more than three active pixels, it undergoes a thinning process specific to that template. The algorithm then proceeds to match the best-fitting template from a set of 12 templates, as illustrated in **Error! Reference source not found.**.

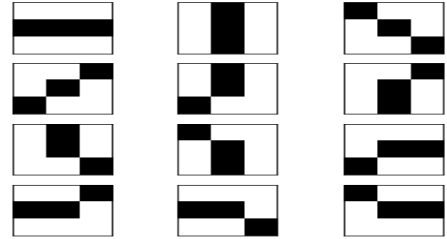

Figure 4 12 different orientations of 3X3 mask used for the thinning process

When a match is found, the algorithm replaces the 3x3 neighbourhood with the chosen template. This adaptive thinning approach ensures that the resulting strokes are consistently one pixel wide.

### iv. FEATURE EXTRACTION AND FEATURE VECTOR GENERATION

This section describes the proposed feature extraction methodology with consideration as each stroke has no more than two endpoints. Nine features have been identified, including endpoints, 4–line elements (vertical, horizontal, slant line with positive and negative slope), and 4 – curved elements (left flat, left deep, right flat, right flat) with the help of 16 different templates which is describe in [55]. Here template template-matching approach has been used to detect the nine other elements. A given image finds a 3x3 neighbourhood of each active pixel. Further, it finds the best match feature template by matching the neighbourhood with nine elements, and a unique feature code is assigned to each pixel based on the template given in figures Figure 5.

The feature extraction algorithm converts a feature code matrix into a 1x25 numerical feature vector. This vector is created by dividing the feature code matrix into 25 blocks of 6x6, normalizing each block, and generating one number per block. These 25 shape numbers serve as stroke features.

To determine the block number, the algorithm identifies the most frequently occurring feature code in each block. To account for endpoints in strokes (limited to two), blocks with endpoints are labelled with the endpoint's feature code (i.e., 10). Additionally, if a block's most frequent code is zero and there are fewer than two non-zero elements, it is set to zero. This accounts for the most common non-zero feature code in the block for other cases. In the classification context, the feature vector data is represented by the first 25 columns of the matrix, while the class label is represented by the last column.

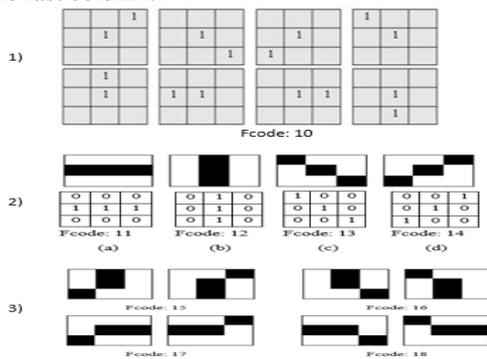

Figure 5 1) template for detecting endpoints with its feature code 2) 3x3 template for line element with its fcode a) horizontal line b) vertical line c) right slant d) left slant 3) 3x3 template for curve element with its features code.

Table 3 Class-wise average accuracy achieved for different classifiers

| Class Label | Decision Tree | Support Vector Machine | KNN | Gradient Boost | Logistic Regression | Naïve Bayes | Random Forest |
|---|---|---|---|---|---|---|---|
| Class 1 | 1 | 1 | 1 | 0.98 | 0.98 | 1 | 0.96 |
| Class 2 | 0.94 | 0.9 | 0.88 | 0.96 | 0.96 | 0.87 | 0.98 |
| Class 3 | 1 | 1 | 0.98 | 1 | 1 | 0.98 | 1 |
| Class 4 | 1 | 1 | 0.96 | 1 | 1 | 1 | 1 |
| Class 5 | 0.92 | 0.9 | 0.94 | 0.96 | 0.96 | 0.86 | 0.96 |
| Class 6 | 0.98 | 1 | 1 | 0.98 | 0.98 | 0.96 | 0.98 |
| Accuracy | **0.97** | **0.97** | **0.96** | **0.98** | **0.98** | **0.95** | **0.98** |

## v. STROKES CLASSIFICATION

The effectiveness of the feature vectors was evaluated through a series of experiments. Several classifiers, such as Decision Tree, Support Vector Machine, KNN, Gradient Boost, Logistic Regression, Naive Bayes, and Random Forest, were implemented to assess the performance of the feature vectors.

The results obtained from each classifier were promising. These classifiers demonstrated the capability to effectively classify the strokes based on the extracted features. The accuracy and performance of the classifiers were encouraging, indicating the potential of our approach for stroke classification in character

Table 2 Testing and verification accuracy for stroke classification with different classifiers

| | Decision Tree | Support Vector Machine | KNN | Gradient Boost | Logistic Regression | Naïve Bayes | Random Forest |
|---|---|---|---|---|---|---|---|
| Testing Accuracy | 0.97 | 0.97 | 0.96 | 0.98 | 0.98 | 0.95 | 0.98 |
| Validation Accuracy | 0.83 | 0.9 | 0.88 | 0.9 | 0.88 | 0.92 | 0.92 |

generation.

In the stroke classification experiment, a dataset comprising 600 different samples was utilized. Each sample contained a feature vector of size 26, consisting of 25 feature data values and 1 class label. To ensure a true evaluation, the dataset was divided into training and testing samples for each class.

Table 3 shows the class-wise average accuracy achieved for different classifiers, including the accuracy for each class and the overall accuracy.

Random Forest, Gradient Boost, Logistic Regression, and Decision Tree classifiers achieve high accuracy across all classes, with class-wise average accuracy ranging from 0.92 to 1. These classifiers demonstrate strong performance in stroke classification. The overall accuracy of the classifiers ranges from 0.95 to 0.98, demonstrating their effectiveness in stroke classification.

Figure 6 represents the accuracy achieved by different classifiers during the testing stage. The results show that Gradient Boost, Logistic Regression, and Random Forest classifiers perform well with an accuracy of 98%. This high accuracy suggests that the proposed features are effective in capturing the relevant information for stroke classification.

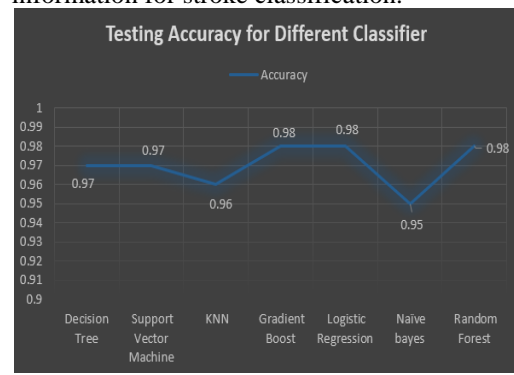

Figure 6 Stroke classification accuracy achieved on different classifiers

Based on the experiments conducted, it is concluded that the structural feature-based feature vectors utilized in this study are considered the finest features. This implies that the chosen features successfully capture the distinctive characteristics of stroke patterns, leading to accurate classification results.

After the completion of the training and testing stages, the model was verified using a separate set of samples. A total of 60 samples were utilized for verification, with 10 samples assigned to each class. The results of the verification process are presented in **Error! Reference source not found.**.

Table 2 presents the testing and verification accuracy for stroke classification using different classifiers. The testing accuracy represents the performance of the classifiers on a separate testing dataset, while the validation accuracy corresponds to the accuracy achieved during the verification stage using a subset of samples.

*Table 4* Class-Wise verification accuracy on different classifiers

|  | Decision Tree | Support Vector Machine | KNN | Gradient Boost | Logistic Regression | Naïve Bayes | Random Forest |
|---|---|---|---|---|---|---|---|
| **Class 1** | 1 | 1 | 1 | 1 | 1 | 1 | 1 |
| **Class 2** | 0.8 | 0.77 | 0.8 | 0.87 | 0.83 | 0.95 | 0.91 |
| **Class 3** | 0.95 | 0.95 | 1 | 0.8 | 0.95 | 0.95 | 0.91 |
| **Class 4** | 0.9 | 1 | 1 | 1 | 0.95 | 1 | 1 |
| **Class 5** | 0.6 | 0.95 | 0.74 | 0.95 | 0.76 | 0.83 | 0.9 |
| **Class 6** | 0.75 | 0.75 | 0.75 | 0.75 | 0.82 | 0.75 | 0.75 |
| **Accuracy** | **0.83** | **0.9** | **0.88** | **0.9** | **0.88** | **0.92** | **0.92** |

The Random Forest classifier consistently achieves a high accuracy of 0.98 in both the testing and validation phases, indicating its strong performance. The Gradient Boost and Logistic Regression classifiers also demonstrate notable accuracy, with testing accuracies of 0.98 and validation accuracies of 0.9, respectively.

On the other hand, the Naïve Bayes classifier exhibits slightly lower accuracy compared to the other models, with a testing accuracy of 0.95 and a validation accuracy of 0.92. The Decision Tree, Support Vector Machine, and KNN classifiers perform well with testing accuracies ranging from 0.96 to 0.97.

Overall, these results indicate that the Random Forest classifier consistently achieves high accuracy and is considered the best model for stroke classification in this context.

### vi. CHARACTER GENERATION

In the character generation stage of the HGFG system, the goal is to create the complete glyph of a Gujarati character based on the strokes extracted and labelled from the user's writing.

The process of character generation begins by selecting the appropriate strokes for the given character. These strokes are then placed and aligned according to the predefined rules. The positioning of each stroke is determined by its specific role within the character and its relationship to other strokes. Once the strokes are positioned, they are concatenated or joined together to form the complete glyph of the character. Figure 7 illustrates the concept of stroke concatenation and character Generation in this stage.

In our current study, we have simplified the character generation process for the Gujarati script. Users only need to write three characters: ક, ખ, and ગ. From these three characters, we have extracted a total of six strokes. By combining these strokes according to predefined rules, we can generate twenty-three additional consonants.

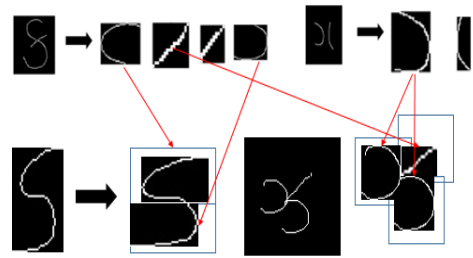

Figure 7 Concept of character generation of Gujarati character from the strokes extracted from user writing

To facilitate this character generation, we have developed Algorithm, which outlines the complete method. The algorithm takes as input the name of the Gujarati character that needs to be generated. It then identifies the specific strokes required to construct that character. Detailed stroke information such as height, width, joining point, position, size, appearance, and occurrence is collected from a prepared ruleset. Finally, using this information, the algorithm generates the glyph of the desired character.

Figure 8 shows a collection of twenty-three characters generated from the three input characters. These characters are the outcome of the glyph synthesis process and demonstrate the diversity of consonant characters that can be generated based on the predefined rules and input strokes.

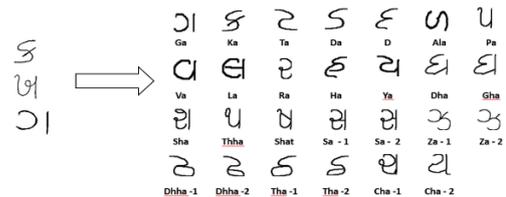

Figure 8 Generated Gujarati characters from strokes extracted from characters ક, ખ, and ગ

### vii. FONT GENERATION

After generating the character as an image, the next step involves converting the image into an outline image, essentially creating a computer font. This conversion is automated using the Font Forge tool[56], an open-source program that enables the creation and modification of fonts in various standard formats.

Successfully, 30 different font files were generated using the FontForge tool by utilizing the generated samples from specific users. Figure 9 showcases a demonstration of some words written in the user's generated font. These font files serve as valuable resources for various applications, including typography, document design, and language processing.

Figure 9 A demonstration of the generated user's handwritten Gujarati font

## 4. RESULTS AND DISCUSSION

In this section, we will focus on evaluating the HGFG framework as a whole. This evaluation includes an assessment of the successful generation of Gujarati characters and how closely they resemble the ground truth. We will also identify areas that may require further improvement. Analyzing the results of this phase provides valuable insights into the effectiveness of the HGFG process in generating handwritten Gujarati fonts.

### a. EVALUATION OF THE HANDWRITTEN GUJARATI FONT GENERATION (HGFG) FRAMEWORK

The evaluation of our HGFG framework involved assessing the quality and accuracy of the generated fonts. Two different evaluation methods were employed to measure the performance of the framework and determine its effectiveness: Subjective Evaluation and Objective Evaluation. These metrics were selected to provide a comprehensive evaluation of the model's performance.

The subjective evaluation involved conducting user studies, where participants provided feedback on the quality and visual appeal of the synthesized images and fonts. Their opinions helped assess the overall performance and user perception of the generated content. Objective metrics were also employed to quantify specific aspects of the generated fonts, such as stroke accuracy, curvature, and overall similarity to the original handwritten characters.

Furthermore, objective metrics were utilized to evaluate the performance of the synthesized system. One such metric involved assessing the impact of the generated fonts on the performance of a Gujarati character recognizer system. The recognition accuracy of the recognizer system when using the synthesized fonts was analyzed, providing quantifiable insights into the effectiveness and compatibility of the generated fonts in practical applications.

Evaluation by User Study

To assess the authenticity of the generated handwritten Gujarati samples, a subjective study involved 30 writers. They rated the resemblance of the generated characters to the original writing. A Word file was provided, with options "Y" for a match, "P" for partial, and "N" for no match. This study gathered qualitative feedback, complementing objective metrics in evaluating the HGFG framework's effectiveness. Ratings were statistically analyzed, providing valuable insights into the quality of the generated characters.

Figure 10 illustrates the character-wise generation accuracy of the HGFG system in chart form. The chart displays the accuracy of each generated consonant. Upon observation, it can be concluded that the characters ક and ગ achieved the highest accuracy of 100%. On the other hand, the character ય had the lowest generation accuracy of 70%.

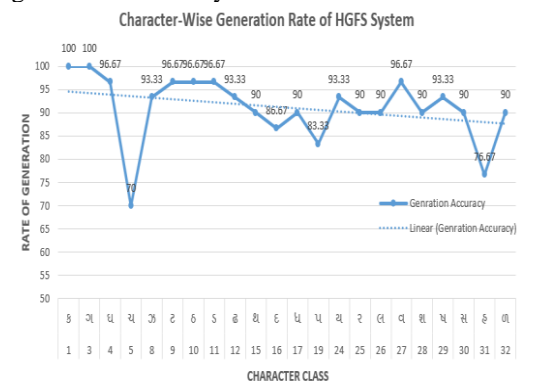

Figure 10 Visual representation of character-wise generation accuracy of HGFG system.

Table 5 Character-wise generation accuracy of the HGFG system

| Character Class | Characters | Generation | Character Class | Characters | Generation |
|---|---|---|---|---|---|

|   |   | Accuracy |    |   | Accuracy |
|---|---|----------|----|---|----------|
| 1 | ક | 100 | 17 | ધ | 90 |
| 3 | ગ | 100 | 19 | પ | 83.33 |
| 4 | ઘ | 96.67 | 24 | ય | 93.33 |
| 5 | ચ | 70 | 25 | ર | 90 |
| 8 | ઝ | 93.33 | 26 | લ | 90 |
| 9 | ટ | 96.67 | 27 | વ | 96.67 |
| 10 | ઠ | 96.67 | 28 | શ | 90 |
| 11 | ડ | 96.67 | 29 | ષ | 93.33 |
| 12 | ઢ | 93.33 | 30 | સ | 90 |
| 15 | થ | 90 | 31 | હ | 76.67 |
| 16 | દ | 86.67 | 32 | ળ | 90 |

Table 5 provides further details on the character-wise generation accuracy of the HGFG system. It includes the character class, corresponding characters, and their respective generation accuracy. The table reveals that for eleven characters, an accuracy of over 90% was achieved.

The subjective measure by user study confirms the effectiveness of the HGFG system and provides valuable insights from real users. This user-centric approach ensures that the generated handwritten Gujarati characters meet user expectations and drive further improvements. The study results will guide future research and development, aiming for higher accuracy and more realistic outcomes.

### i. OBJECTIVE EVALUATION OF HGFG THROUGH RECOGNITION SYSTEM

The objective evaluation of the HGFG system involved using an existing recognition system developed by a [57]. The system processed a dataset of these characters, comparing its recognition results to ground truth labels. The accuracy was calculated as the percentage of correctly recognized characters. Testing encompassed 30 samples of each character class. Table 6 displays the evaluation results, revealing recognition accuracy and misclassifications. The associated characters with their deviation in accuracy are presented in Figure 11.

Table 6 Recognition rate of generated characters of HGFG system

| Characters | Objective Evaluation by Recognition System | Characters | Objective Evaluation by Recognition System |
|---|---|---|---|
| ક | 93.3 | ધ | 80 |
| ગ | 96.7 | પ | 70 |
| ઘ | 83.3 | ય | 70 |
| ચ | 66.7 | ર | 90 |
| ઝ | 80 | લ | 83.3 |
| ટ | 93.3 | વ | 80 |
| ઠ | 83.3 | શ | 86.7 |
| ડ | 90 | ષ | 76.7 |
| ઢ | 80 | સ | 80 |
| થ | 73.3 | હ | 70 |
| દ | 70 | ળ | 96.7 |

The study reveals varying recognition rates for different characters. Characters like ગ, ક, ટ, and ળ achieve high accuracy (93.3% to 96.7%), indicating successful classification by the recognition system.

Conversely, characters like ચ, પ, ષ, and થ exhibit lower recognition rates (56.7% to 83.3%), posing challenges for the system with more misclassifications.

The evaluation results indicate that the HGFG system is effective in generating characters that can be recognized with reasonably good accuracy for certain specific characters. These accurately recognized characters can be considered successful outputs of the HGFG system.

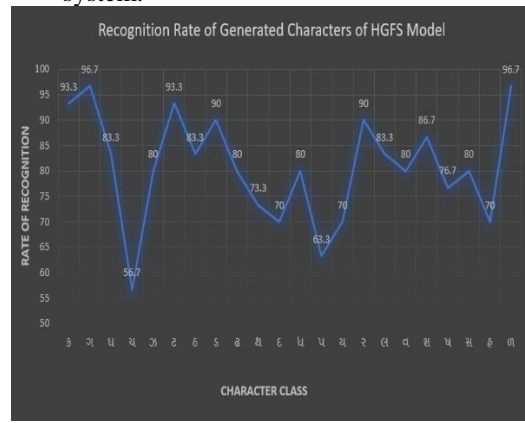

Figure 11 Recognition rate of generated characters of HGFG system.

### ii. COMPARISON OF HGFG MODEL EVALUATION: RECOGNITION SYSTEM VS USER INVOLVEMENT

In this section, a comparison is made between the evaluation of the HGFG model using two different methods. Figure 12 provides a visual representation of this comparison.

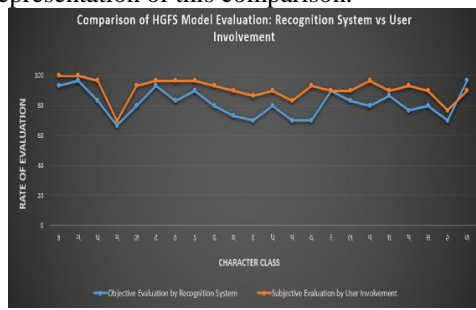

Figure 12 Comparison of HGFG model evaluation: recognition system vs. user involvement

Characters like 'ક' (Ka), 'ટ' (Ta), 'ઠ' (Tha), and 'ડ' (Da) and 'ગ' (Ga) achieved accuracies of 93.3% and 96.7% in both the recognition system and user evaluations. For example, 'ચ' (Cha) had a 56.7% accuracy in the recognition system but was rated at 70% accuracy by users. Overall, the comparison highlights the importance of considering both objective and subjective evaluations in assessing the performance of the HGFG model.

## 5.  CONCLUSION

The research aimed to create handwritten Gujarati fonts using an approach called HGFG, involving two key phases: Learning and Generation. In the Learning phase, the study examined Gujarati consonant glyphs to understand their visual characteristics. A ruleset with six essential strokes was developed for stroke-based generation. The Generation phase covered preprocessing, segmentation, classification, character and font creation, and verification. Classification models achieved high accuracy (95%-98%) for stroke classification.

The synthesized handwritten Gujarati fonts generated by the HGFG framework closely resemble the ground truth. The evaluation of the framework includes both subjective and objective metrics. Subjective evaluation by user studies provided valuable feedback on the quality and visual appeal of the synthesized characters and fonts. The subjective evaluation indicated high accuracy ranging 90-96.67 and lower accuracy (70%) in characters like 'ચ'.

Objective evaluation using a recognition system further confirmed the effectiveness of the synthesized characters and achieved high recognition rates, with accuracy ranging from 93.3% to 96.7%. This indicates that the recognition system was successful in correctly classifying the majority of these characters Overall, the HGFG framework shows promising results in generating accurate and visually appealing handwritten Gujarati fonts.

Future work should focus on refining and further developing the framework, addressing observed flaws in characters including the presence of extra noise, incorrect concatenation of components, and sharp curves that are not smooth. Additionally, expanding the scope of the study to include the generation of complete Gujarati fonts, encompassing vowels, digits, diacritic marks, and special symbols, would be a valuable direction for future research. The findings of this research serve as a solid foundation for future advancements in Gujarati font synthesis, with potential applications in typography and design.